\documentclass[10pt, journal]{IEEEtran}
\usepackage{amsmath, amsfonts, amssymb}
\usepackage{multirow, makecell, array}
\usepackage[caption=false, font=normalsize, labelfont=sf, textfont=sf]{subfig}
\usepackage{textcomp}
\usepackage{stfloats}
\usepackage{url}
\usepackage{verbatim}
\usepackage{graphicx}
\usepackage{inputenc}
\usepackage{amsthm}
\usepackage{booktabs}
\usepackage{algorithm}
\usepackage{algorithmic}
\usepackage{adjustbox}
\usepackage[table]{xcolor}
\definecolor{ours}{rgb}{0.98, 0.85, 0.87}

\newcommand{\ie}{\textit{i}.\textit{e}.}

\newcommand{\affiliation}[1]{\textsuperscript{#1}}
\newcommand{\equalcontribution}{\textsuperscript{†}}
\newcommand{\corresponding}{\textsuperscript{*}}

\begin{document}

\title{R$^2$MoE: Redundancy-Removal Mixture of Experts for Lifelong Concept Learning}

\author{
Xiaohan Guo\equalcontribution, Yusong Cai\equalcontribution, Zejia Liu\equalcontribution, Zhengning Wang\affiliation{a},~\IEEEmembership{Member,~IEEE}, 
Lili Pan\corresponding\affiliation{a},~\IEEEmembership{Member,~IEEE}, 
Hongliang Li\affiliation{a},~\IEEEmembership{Senior Member,~IEEE}%
\thanks{Xiaohan Guo, Yusong Cai, Zejia Liu, Zhengning Wang, Lili Pan, and Hongliang Li are with the School of Information and Communication Engineering, University of Electronic Science and Technology of China, Chengdu, 611731.}%
\thanks{This work is partially supported by the National Natural Science Foundation of China (No.~62171111).}%
\thanks{\equalcontribution denotes equal contribution.}%
\thanks{\corresponding denotes corresponding author.}%
}

\markboth{Journal of \LaTeX\ Class Files,~Vol.~14, No.~8, August~2021}{Guo \MakeLowercase{\textit{et al.}}: R$^2$MoE for Lifelong Concept Learning}

\IEEEpubid{0000--0000/00\$00.00~\copyright~2021 IEEE}

\maketitle

\begin{abstract}
Enabling large-scale generative models to continuously learn new visual concepts is essential for personalizing pre-trained models to meet individual user preferences.
Existing approaches for continual visual concept learning are constrained by two fundamental challenges: catastrophic forgetting and parameter expansion.
In this paper, we propose Redundancy-Removal Mixture of Experts (R$^2$MoE), a parameter-efficient framework for lifelong visual concept learning that effectively learns new concepts while incurring minimal parameter overhead.
Our framework includes three key innovative contributions:
First, we propose a mixture-of-experts framework with a routing distillation mechanism that enables experts to acquire concept-specific knowledge while preserving the gating network's routing capability, thereby effectively mitigating catastrophic forgetting.
Second, we propose a strategy for eliminating redundant layer-wise experts that reduces the number of expert parameters by fully utilizing previously learned experts.
Third, we employ a hierarchical local attention-guided inference approach to mitigate interference between generated visual concepts.
Extensive experiments have demonstrated that our method generates images with superior conceptual fidelity compared to the state-of-the-art (SOTA) method, achieving an impressive 87.8\% reduction in forgetting rates and 63.3\% fewer parameters on the CustomConcept 101 dataset.
Our code is available at
{https://github.com/learninginvision/R2MoE}
\end{abstract}


\section{Introduction}
\noindent \IEEEPARstart{L}arge-scale generative models such as Stable Diffusion and
Imagen~\cite{sd,sdxl,sd3,Imagen}, based on probabilistic diffusion theory~\cite{ddpm}, have demonstrated remarkable performance in text-to-image (T2I) generation. 
These models excel at producing visual outputs with exceptional fidelity and rich semantic diversity in response to textual prompts. 
However, despite their strengths in open-domain generation, as general-purpose models they often face challenges when generating personalized concepts.
This limitation has driven research into personalized generation methods~\cite{textureinversion,dreambooth,oft,Personalized-Residuals,xu2024sgdm,zhang2024two,chen2025videodreamer}, which enable efficient adaptation of pre-trained T2I models using only a few reference images. Such approaches allow these models to create customized content while preserving their broad generative capabilities.

\begin{figure}[t]
    \centering  
    \includegraphics[width=1.0\linewidth]{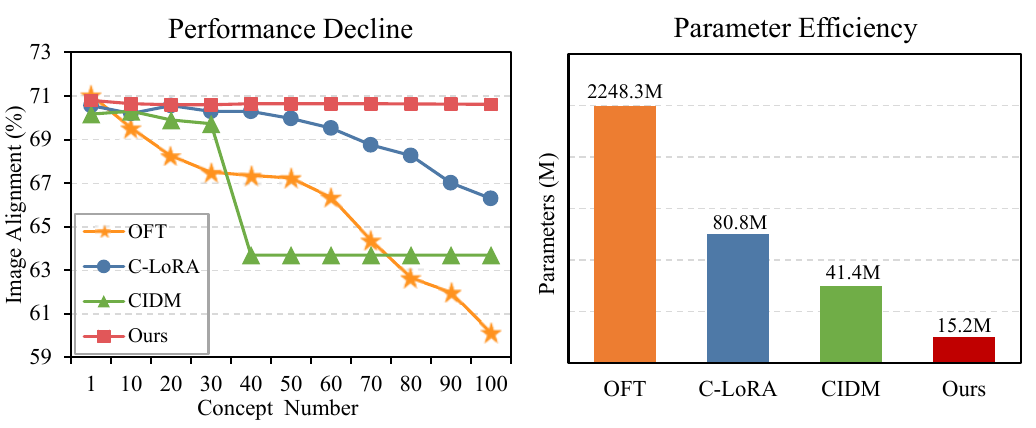}
    \vspace{-15pt}
    \caption{
    \textbf{R$^2$MoE vs. state-of-the-Art methods on CustomConcept101 Dataset.}
    \textbf{Left:} R$^2$MoE achieves superior concept retention with minimal degradation in image alignment, significantly outperforming C-LoRA and CIDM by $4.3\%$ and $10.4\%$, respectively.
    \textbf{Right:} our approach achieves the lowest parameter overhead, reducing parameter count by $81.2\%$ and $63.3\%$ compared to C-LoRA and CIDM,
respectively.
   }
    \label{fig:motivation}
    \vspace{-10pt}
\end{figure}


Personalizing pre-trained large-scale generative models typically involves the continual integration of new concepts, either by updating key parameters~\cite{customdiffusion,han2023svdiff,animediff} or by learning Low-Rank Adaptations (LoRA) tailored to the pretrained model~\cite{mixofshow,cvpr2024orthogonal,Multi-LoRA,zhang2025multi}. 
However, such adaptations often lead to catastrophic forgetting, substantially impairing the model’s ability to generate both previously learned concepts and those acquired during pre-training. 
To address this challenge, recent studies~\cite{clora,l2dm,CIDM,guo2025conceptguard} have approached the problem from a continual learning perspective~\cite{mccloskey1989catastrophic}, making notable progress in mitigating forgetting while enabling the acquisition of new concepts.
For example, C-LoRA~\cite{clora} introduces a regularized LoRA module that mitigates catastrophic forgetting by constraining updates to previously adapted parameters from earlier tasks.
CIDM~\cite{CIDM} captures concept-specific knowledge through distinct LoRA modules and leverages an elastic distillation mechanism to adaptively weight them based on the semantic content of input prompts during inference. 
However, as the number of learned concepts grows, the performance of these methods progressively declines (Fig.\ref{fig:motivation}, left), indicating that catastrophic forgetting remains a persistent challenge. 
Furthermore, these approaches do not effectively mitigate the issue of increasing parameter overhead in continual concept learning (Fig.\ref{fig:motivation}, right), ultimately constraining the model’s scalability and its capacity for lifelong learning.\IEEEpubidadjcol

This work formalizes Lifelong Concept Learning (LCL) as a learning paradigm for continuously integrating new concepts into large-scale generative models while maintaining minimal interference with previously acquired knowledge.
This learning paradigm confronts two fundamental
challenges:
\textit{catastrophic forgetting} and \textit{parameter growth}.

\begin{figure*}[t!]
    \centering  
    \includegraphics[width=0.9\linewidth]{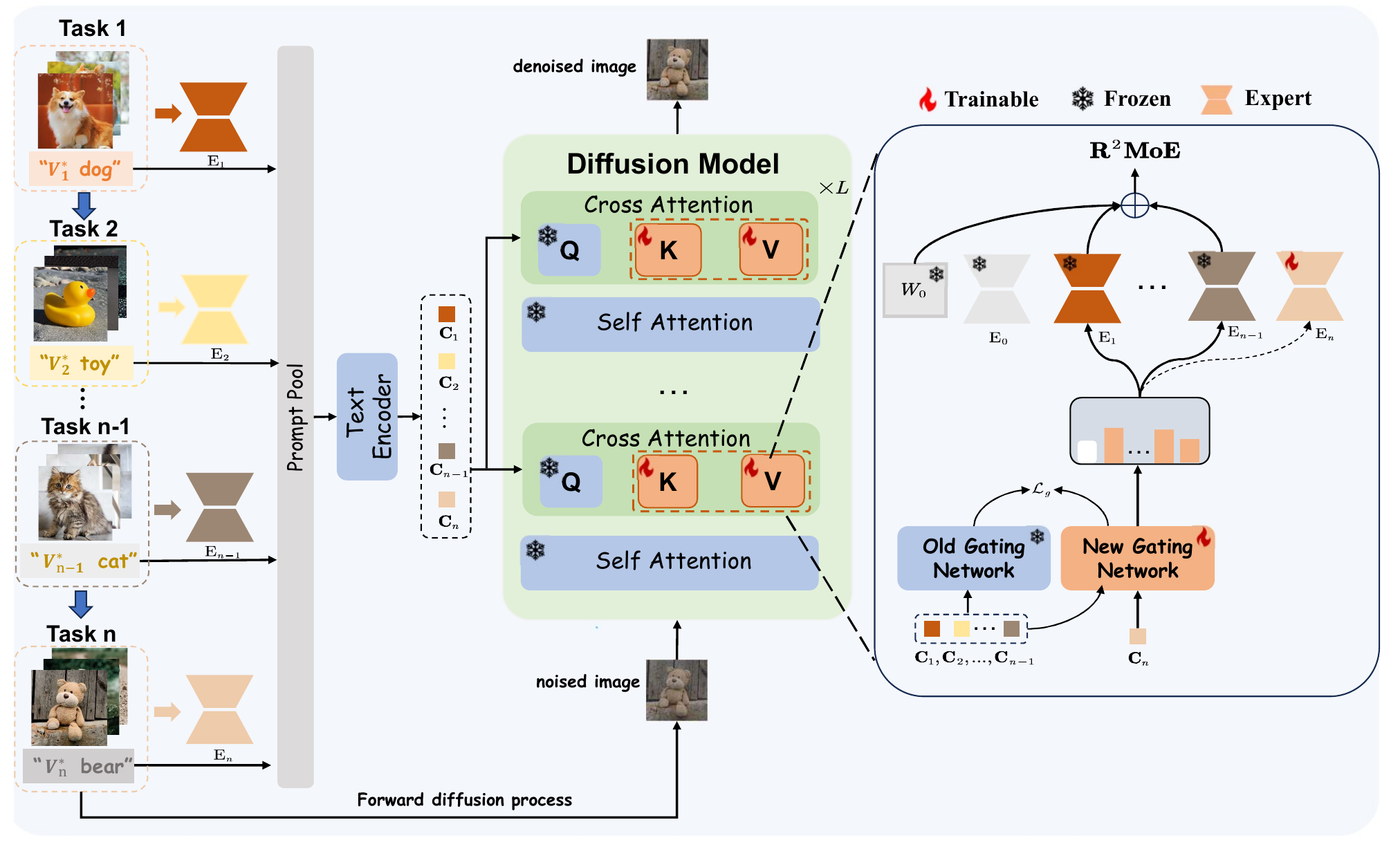}
    \vspace{-10pt}
    \caption{\textbf{Train paradigm of the proposed R$^2$MoE framework.}
    The fig illustrates a continual learning process across multiple tasks (1 through n), where each task introduces a distinct visual concept (e.g., "dog", "toy", "cat", "bear").
    When learning the $n$-th task, the unique identifiers and category information of previously learned concepts, along with the current task's prompt, are fed into the text encoder to obtain text embeddings.
    The text embeddings of previous concepts are used to distill knowledge from the old gating network to the new gating network, ensuring that previously learned routing capabilities are not forgotten.
    Simultaneously, the text embedding of the current task guides the expert selection process, utilizing a combination of new and old experts to facilitate learning of the new task.
    After each task is completed, the corresponding expert is frozen to preserve the acquired knowledge.}
    \label{fig:overview}
    \vspace{-10pt}
\end{figure*}

To address these problems, we propose Redundancy-Removal Mixture of Experts (R$^2$MoE), a novel personalized mixture of experts framework that integrates a routing distillation mechanism (as shown in Fig.\ref{fig:overview}).
R$^2$MoE leverages the sparse routing property of MoE mitigates the disruptive impact of learning new concepts on the model parameters.
The routing distillation mechanism distills the gating network using concept-specific tokens, enabling the model to retain the ability to route previously learned concepts to their corresponding experts and mitigates catastrophic forgetting.
Moreover, since the gating network merely selects routing paths or assigns expert weights, this approach enables efficient knowledge transfer without compromising expert specialization, thereby enhancing the model's generalization capabilities.
To reduce parameter overhead in lifelong concept learning, we propose a strategy to eliminate redundant layer-wise experts.
For this, we validate the existence of redundant layer-wise experts by reusing previously learned experts.
Then, we define an importance metric to evaluate the significance of each layer-wise expert for the current task, and prune experts with low importance.

Finally, we propose a novel hierarchical local attention-guided inference approach, which facilitates expert collaboration.
This strategy first employs a LLM~\cite{GPT4} to automatically generate region prompts and bounding boxes, guiding the spatial layout of generated content.
Subsequently, a semantic segmentation model (SAM)~\cite{sam} refines these regions to produce more precise attention mask for concepts, thereby effectively mitigating feature entanglement in multi-concept generation.

In summary, the main contributions are as follows:
\begin{itemize}
\item 
We propose R$^2$MoE, a novel mixture
of experts framework with a routing distillation mechanism for lifelong concept learning.
It enables the continual integration of new visual concepts into a pre-trained T2I model while effectively mitigating catastrophic forgetting.

\item To eliminate redundant parameters, we propose a redundant layer-wise experts eliminating strategy that reduces the number of expert parameters by fully leveraging previously learned experts.

\item To mitigate feature entanglement in single/multi-concept generation, we propose a hierarchical local attention-guided inference approach that first employs a LLM to generate image layout information, then utilizes a SAM to obtain region attention masks for different concepts.

\item Our R$^2$MoE is validated to outperform existing state-of-the-art lifelong concept learning approaches.
Notably, R$^2$MoE achieves a CILP-IA of $75.3\%$ on CustomConcept101 dataset, surpassing CIDM~\cite{CIDM} by $10.1\%$ and C-LoRA~\cite{clora} by $2.8\%$.
In addition, our method demonstrates significantly low forgetting rate of only $0.19\%$, compared to $8.56\%$ for CIDM and $1.56\%$ for C-LoRA.
Moreover, our approach achieves the lowest parameter overhead, reducing parameter count by $81.2\%$ and $63.3\%$ compared to C-LoRA and CIDM, respectively. 

\end{itemize}
\section{Related Work}

\subsection{Personalized Large-Scale Generative Model}
To generate \textit{concept-specific} images from textual prompts, numerous studies~\cite{textureinversion,instantbooth,dreambooth,customdiffusion,oft,zhang2024two,xu2024sgdm,zhang2025multi,chen2025videodreamer} have focused on fine-tuning large-scale T2I diffusion models~\cite{sd,sdxl,DALLE-2} using personalized datasets.
Texture Inversion~\cite{textureinversion} first introduced personalized T2I generation by learning a pseudo-word in the text encoder's embedding space, thereby capturing high-level personalized semantics.
DreamBooth~\cite{dreambooth} adopts full model fine-tuning to improve personalization while employing a class-specific prior preservation loss to mitigate \textit{overfitting} and \textit{language drift}.
OFT~\cite{oft} effectively preserves the world knowledge of the pre-trained model by adding orthogonal matrices to learn new concept.
Several studies~\cite{han2023svdiff,mixofshow,Multi-LoRA,cvpr2024orthogonal,animediff} have explored methods for multi-concept generation. 
For example, Custom Diffusion~\cite{customdiffusion} restricts fine-tuning to the key and value projection matrices within cross-attention layers, allowing multi-concept generation through joint training across several concepts.

Despite their effectiveness, these personalized generation methods are typically limited to learning a small number of concepts.
Recent works~\cite{clora,l2dm,CIDM,guo2025conceptguard} have begun to address this limitation by enabling the sequential learning of multiple concepts.
C-LoRA~\cite{clora} continually learns new concepts through self-regularized low-rank matrices~\cite{lora}, effectively preventing forgetting of previously learned concepts.
L2DM~\cite{l2dm} employs sample replay and model distillation to alleviate forgetting.
CIDM~\cite{CIDM} introduces a concept consolidation loss and an elastic weight aggregation module to mitigate the catastrophic forgetting of old personalized concepts, by exploring task-specific/task-shared knowledge and aggregating all low-rank weights of old concepts based on their contributions.
However, these methods face challenges of catastrophic forgetting and degraded generalization in long task sequence.


\subsection{Mixture of Experts}
The Mixture of Experts (MoE)~\cite{moe} paradigm typically consists of a gating network and multiple expert modules, where the gating network dynamically assigns coefficients to each expert based on the input, resulting in a weighted combination of expert outputs.
MoE methods offer substantial advantages in terms of computational efficiency and scalability.
For example, Sparse MoE~\cite{sparsemoe} activates only a subset of model parameters during training and inference, significantly reducing computational overhead while maintaining strong performance.
Beyond efficiency, MoE frameworks also improve task-specific performance by leveraging the complementary strengths of multiple experts.
For instance, MoA~\cite{MOA} employs an explicit routing strategy during training to extract diverse knowledge from experts, boosting single-task performance.
MoLE~\cite{MOLE} combines multiple pre-trained LoRA modules using learnable hierarchical gating functions, enabling flexible and efficient expert composition.
Given these properties, MoE models are naturally well-suited to continual learning scenarios.
Expert-Gate~\cite{Expert-gate} introduces a gating mechanism using autoencoders to encode task representations and route test samples to the most appropriate experts.
Lifelong-MoE~\cite{Lifelong-MoE} dynamically expands model capacity by adding new experts through a regularized training process.
MoE-Adapters~\cite{boosting-MOE} insert MoE-based adapters into a pre-trained CLIP model, enabling incremental adaptation to new tasks while maintaining flexibility and task-specific specialization.
LiveEdit~\cite{liveedit} retains expert modules corresponding to each edit instance during training. 
At inference time, it computes expert coefficients based on the semantic and visual features of the input, thereby enabling lifelong vision-language model editing.

However, a major limitation of existing MoE-based lifelong learning approaches lies in their growing memory footprint, as new experts are added for each task or concept, the overall model size increases linearly.
To address this challenge, we propose a lightweight and scalable integration of the MoE framework into lifelong concept learning. 

\section{Preliminaries}
\subsection{Text-to-Image Diffusion Models}
Latent diffusion models (LDMs)~\cite{sd} alleviate the computational burdens of traditional diffusion models by operating in a learned latent space rather than directly in the pixel space. 
LDMs typically consist of two key components.
First, an autoencoder, pre-trained on large-scale image datasets, learns a low-dimensional yet semantically rich latent representation. This representation preserves the essential spatial and semantic structure of the original image while significantly reducing computational complexity.
Specifically, the encoder $\mathcal{E}$ maps an image $\mathcal{x}$ into a latent feature $\mathbf{z}=\mathcal{E}(x)$, and the decoder $\mathcal{D}$ reconstructs the image from the latent representation $\tilde{x}=\mathcal{D}(\mathbf{z})=\mathcal{D}(\mathcal{E}(x))$.
The second component is a diffusion model that operates directly in the latent space to learn the generative process.
In this stage, various forms of conditioning information can be integrated to guide generation. For text-to-image tasks, a text encoder $\psi_{\boldsymbol{\phi}}(\cdot)$ maps the conditional text prompt $c$ into a latent representation $\psi_{\boldsymbol{\phi}}({c})$, which modulates the image generation process. LDMs are trained using a denoising objective defined as:
\begin{equation}
\mathcal{L}_{\text{LDM}}(\boldsymbol{\theta}) = \mathbb{E}_{\mathcal{E}(x),c,\mathbf{\epsilon} \sim \mathcal{N}(\mathbf{0},\mathbf{I}),t}\left[ || \mathbf{\epsilon} - \mathbf{\epsilon}_{\boldsymbol{\theta}}(\mathbf{z}_{t}, t, \psi_{\boldsymbol{\phi}}(c)) ||_{2}^{2} \right],\label{con:ldm}
\end{equation}
where $\mathbf{z}_t$ represents the latent feature at timestep $t$, $\mathbf{\epsilon}_{\boldsymbol{\theta}}$ denotes the U-Net parameterized by $\boldsymbol{\theta}$, and $\psi_{\boldsymbol{\phi}}$ refers to the  conditional CLIP text encoder
parameterized by $\boldsymbol{\phi}$.

\subsection{Continual Learning Formulation}
In this paper, we define a sequence of tasks $\mathcal{N}=\{\mathcal{N}_n\}_{n=1}^N$, each associated with a corresponding dataset $\mathcal{D}=\{\mathcal{D}_n\}_{n=1}^N$. 
The $n$-th generative task consists of $|\mathcal{D}_n|$ input pairs, where each pair comprises a conceptual image $x_{n,i}$ and a textual prompt $c_{n}$.
For example, a prompt could be \textit{photo of a $V^{*}_n$ $class_n$}. 
Here, $N$ represents the total number of tasks, and each dataset contains between $3$ and $8$ such pairs.

For each task, the generative model comprises a U-Net $\epsilon_{\boldsymbol{\theta}}(\cdot): \mathbb{R}^{W\times H\times C} \to \mathbb{R}^{W\times H\times C}$ and a text encoder $\psi_{\boldsymbol{\phi}}(\cdot): \mathbb{R}^{L}\to\mathbb{R}^{L\times d}$, where $L$ is the length of the text prompt and $d$ is the dimensionality of each token embedding.
The objective of continual learning in this context is to train a U-Net $\epsilon_{\boldsymbol{\theta}}(\cdot)$ and a text encoder $\psi_{\boldsymbol{\phi}}(\cdot)$ such that the model is capable of generating personalized images for the current task while retaining knowledge from previously learned tasks.

\section{Methodology}
In this section, we introduce our proposed methodology for lifelong concept learning.
Firstly, we provide a description of our R$^2$MoE framework and route distillation mechanism.
Consequently, we present the explanation about the redundant expert layer eliminating strategy.
At last, we introduce the hierarchical local attention-guided inference method.  
An overview of the proposed train paradigm is illustrated in Fig.\ref{fig:overview} and the
inference paradigm is illustrated in Fig.\ref{fig:mul-gen-method}.

\subsection{Lifelong Concept Learning with Mixture of Experts}
\label{sec:S4A}
The Mixture of Experts (MoE) framework offers inherent advantages for continual learning, as it enables modular and parameter-efficient adaptation while isolating concept-specific knowledge to mitigate interference and forgetting~\cite{boosting-MOE,liveedit,coin}. 
To achieve lifelong concept learning, we propose a scalable expert module that dynamically routes inputs to the most relevant expert through a learnable gating network, accommodating an expanding set of concepts without disrupting previously acquired knowledge.

Following prior works~\cite{customdiffusion,clora}, we built our mixture of LoRA experts framework within the $\textbf{W}^K,\textbf{W}^V$ matrices of the cross-attention layers.
Since each matrix operates in the same manner, we illustrate our method using a single matrix for simplicity, omitting layer-specific and $K,V$ notation.
Assuming the text prompt for the $n$-th task is denoted as $c_n$, for example, \textit{photo of a $V^*$ dog}, and $\psi_{\boldsymbol{\phi}}$ is the text encoder parameterized by $\boldsymbol{\phi}$, then the corresponding text embedding is given by $\mathbf{C}_n=\psi_{\boldsymbol{\phi}}(c_n)\in \mathbb{R}^{L\times d_{in}}$, where $L$ is number of the tokens, and $d_{in}$ denotes the embedding dimension of each token. 
We extend the pre-trained LDM parameters $\boldsymbol{\theta}$ with a set of expert modules to enable continual learning. 
A key challenge arises at the early stage of training: since only one expert is active for the first task, the gating network receives no gradient signal and thus fails to learn meaningful routing. 
To address this, we introduce an auxiliary expert $\mathbf{E}_0$, which is initialized at the beginning of training and kept frozen throughout. 
This encourages the gating network to explore other experts by distributing coefficients beyond the frozen expert, thereby improving routing dynamics and alleviating early-stage underfitting.
The parameters of the MoE block are typically formulated as follows:
\begin{equation}
\textbf{W}_n=\textbf{W}_0+ \sum_{i=0}^ng_{n,i}\cdot\mathbf{E}_i,\label{moe}
\end{equation}
where ${\mathbf{W}_n \in \mathbb{R}^{{d_{in}} \times d_{out} }}$ denotes the adapted weight combination used for task $n$, and $\mathbf{W}_0 \in \mathbb{R}^{{d_{in}} \times d_{out}}$ denote a fixed pretrained parameter, which is part of the overall parameter set $\boldsymbol{\theta}$.
The ${\mathbf{E}_{i}\in \mathbb{R}^{{d_{in}} \times d_{out} }}$ is a learnable task-adaptive extension to $\mathbf{W}_0$, and $g_{n,i}$ is the gating coefficient of expert $\mathbf{E}_i$ for task $n$.
The expert coefficients are generated by a gating function $\mathbf{g}_n(\cdot)$ defined as:
\begin{equation}
[g_{n,0}, g_{n,1}, \dots, g_{n,N}]^\top=\mathbf{g}_n({\mathbf{C}_n}) \in \mathbb{R}^{N+1} ,
\end{equation}
where $\mathbf{g}_n(\cdot)\!:\mathbb{R}^{L\times d_{in}} \to \mathbb{R}^{N+1}$ denotes the gating function parameterized by $\mathbf{\Theta}_n$  (we drop $\mathbf{\Theta}_n$ for $\mathbf{g}_n(\cdot, \mathbf{\Theta}_n)$ for simplicity, where $\mathbf{\Theta}_n$ represents the parameters of $n$-th gating network).
It first computes the mean of the text embeddings across the sequence dimension (\ie, averaging over the $L$ tokens), and then passes the result through a trainable MLP layer to produce the expert coefficients.
During training, only the expert parameters ${\mathbf{E}_i}$ and the gating function parameters ${\mathbf{\Theta}_n}$ are optimized via the Eq.~\ref{con:ldm}, while the pre-trained backbone $\mathbf{W}_0$ remains fixed.

To address catastrophic forgetting in LCL, we propose a routing distillation method that enhances the gating network's ability to retain and select experts for prior tasks while adapting to new ones.
Let $\mathbf{g}_{n}(\cdot)$ denote the gating network when learning $n$-th task.
Our method transfers routing knowledge from $\mathbf{g}_{n-1}(\cdot)$  to $\mathbf{g}_{n}(\cdot)$ via knowledge distillation.
Specifically, we store the inputs to the gating network, i.e., the textual embedding of each concept, and define the routing distillation loss as:
\begin{equation}
\mathcal{L}_{\text{g}}(\mathbf{\Theta}_n)=\sum_{\tau=1}^{n-1}\left \| (\mathbf{g}_{n}(\textbf{C}_\tau)-\mathbf{g}_{n-1}(\textbf{C}_{\tau}) \right \|_F^2,
\label{con:OCD}
\end{equation}
where $\mathbf{C}_\tau \in \mathbb{R}^{L\times d_{in}}$ denotes the  concept-specific text embedding corresponding to the ${\tau}$-th concept.
This loss ensures output-space consistency between consecutive gating networks, enabling $\mathbf{g}_{n}(\cdot)$ to inherit and refine $\mathbf{g}_{n-1}(\cdot)$ routing ability without overwriting past knowledge.
Furthermore, this approach avoids imposing constraints directly on the expert modules, thus mitigates the model plasticity deterioration typically observed during sequential LoRA weight fusion~\cite{clora,guo2025conceptguard}.

For each task $n$, the training loss is defined as the combination of generation loss and distillation loss:
\begin{equation}
\begin{aligned}
\mathcal{L}(\boldsymbol{\theta}',\boldsymbol{\phi}) 
&= \mathbb{E}_{\mathcal{E}(x_n),c_n,\mathbf{\epsilon} \sim \mathcal{N}(\mathbf{0},\mathbf{I}),t}
\left[ \left\| \mathbf{\epsilon} - \mathbf{\epsilon}_{\boldsymbol{\theta}'}\big(\mathbf{z}_{t}, t, \psi_{\boldsymbol{\phi}}(c_n)\big) \right\|_{2}^{2} \right] \\
&\quad + \beta\mathcal{L}_{\text{g}}(\mathbf{\Theta}_n),
\label{con:loss}
\end{aligned}
\end{equation}
where $\boldsymbol{\theta}'=(\boldsymbol{\theta},\mathbf{E}_n,\mathbf{\Theta}_n)$
denotes the model parameters during training on the $n$-th task, with $\boldsymbol{\theta}$ being kept frozen and only $\mathbf{E}_n$ and $\mathbf{\Theta}_n$ being updated. 
The parameters $\mathbf{E}_n,\mathbf{\Theta}_n$ are additional parameters introduced into the pre-trained U-Net model and thus can be updated jointly using the LDM loss.
$\boldsymbol\phi$ denotes the  parameters of text encoder and we only fintune the parameters of the token embedding layer.
The scalar $\beta$ is a hyper-parameter that balances the contribution of the routing distillation loss in the overall objective.

\begin{figure}[t!]
    \centering
    \includegraphics[width=1\linewidth]{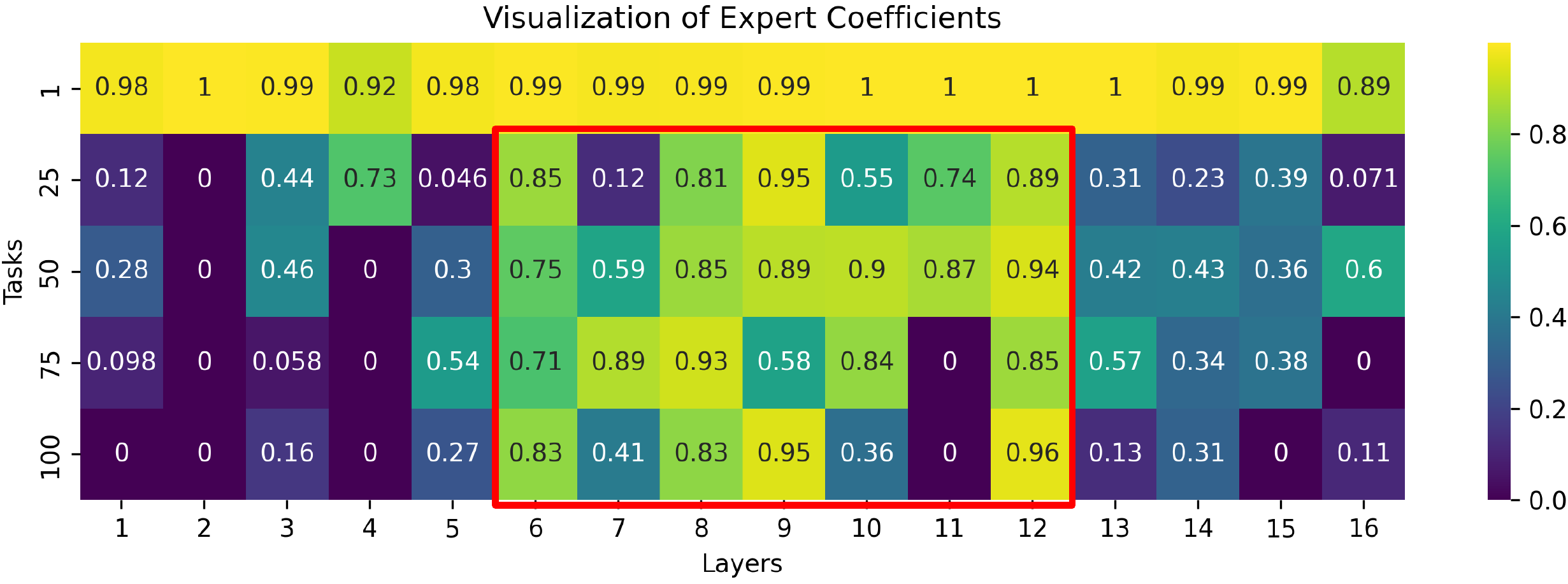}
    \caption{\textbf{Expert Coefficient Magnitudes Across Tasks and Network Layers.}
    As task number increases, experts of middle layers receive higher coefficients (highlighted within the red box), while the remaining experts get smaller coefficients.
    This trend indicates the presence of redundant experts in the model.
    } 
    \label{fig:heatmap}
    \vspace{-5pt}
\end{figure}
\subsection{Selectively Activating Experts}
\label{sec:S4B}

Like Sparse MoE, our framework selectively activates $K$ experts during training. 
To address the gating network’s inherent bias toward previously learned experts, we refine the selection mechanism by choosing $K-1$ experts from the pool of trained experts while enforcing deterministic inclusion of the $0$-{th} expert and the newly added expert. 
The detailed expert selection mechanism and coefficient normalization process are described below:
\begin{equation}
\alpha_{n,i} = \frac{e^{g_{n,i}}}{\sum\limits_{i \in \mathcal{I}_n} e^{g_{n,i}}},
\end{equation}
where $\alpha_{n,i}$ denotes the coefficient assigned to the $i$-th expert for the $n$-th task. $\mathcal{I}_n = [0;\text{TopK}(g_{n,1}, \dots, g_{n,n-1});n] $ denotes the set of selected expert indices, where $\text{TopK}(\cdot)$ selects the $K$ experts with the largest gating coefficients. Notably, during inference, we do not add $n$ to $\mathcal{I}_n$.
\subsection{Eliminating Layer-wise Redundant Experts}
\label{sec:S4C}

MoE architectures often suffer from issues of  knowledge hybridity and knowledge redundancy~\cite{dai2024deepseekmoe}, where each expert acquires non-overlapping and focused knowledge, leading to parameter redundancy across experts. 
To mitigate the problem of knowledge redundancy in lifelong learning scenarios, we propose an effective threshold-based expert pruning strategy.

\begin{figure}[t]
    \centering  
    \includegraphics[width=1\linewidth]{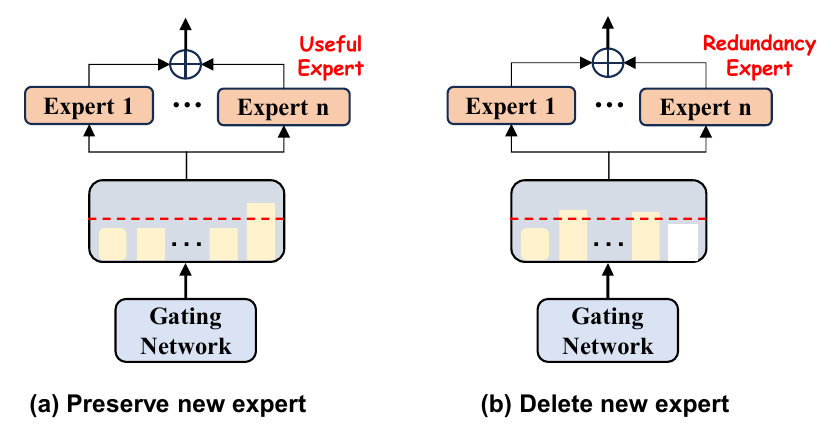}
    \vspace{-10pt}
    \caption{\textbf{The strategy of evaluating expert redundancy.}
    \textbf{Left:} the new expert is retained as its coefficient exceeds the threshold; \textbf{right:} the new expert is deleted as its coefficient falling below the threshold.}
    \label{fig:redundancy_expert}
    \vspace{-10pt}
\end{figure}

Specifically, we ground the learning of new concepts in the weights of both prior experts and pre-trained model, enabling the effective exploitation of existing knowledge to facilitate the acquisition of new concepts. 
By quantitatively analyzing the expert coefficients across different layers of the model, we observe that the gating network consistently assigns smaller coefficients to a subset of experts in the early/late layers as well as a few experts in the intermediate layers (Fig.\ref{fig:heatmap}).
Through~\cite{matte}, early/late layers of diffusion model encode low-level attributes (e.g., color palettes and basic textures) while intermediate layers capture high-level semantics (e.g., object structures and spatial layouts).
This architectural characteristic indicates that certain features (e.g., color schemes) can already be generated by the exist model.
This suggests that, via sequential knowledge transfer, the newly added experts may become redundant, contributing minimally to the overall model capacity.
Based on the aforementioned findings, we propose a simple yet effective approach to quantify the importance of each expert and eliminate unnecessary experts, as illustrated in Fig.\ref{fig:redundancy_expert}.
Specifically, we interpret the expert coefficients $\boldsymbol{\alpha}_n \in \mathbb{R}^{N+1}$ as indicators of expert importance for the $n$-th task.
Upon completion of each task's training, new experts are retained only if their coefficient allocation by the gating network exceeds threshold $p$; otherwise, they are eliminated.
The parameter of the MoE block can be expressed as:
\begin{equation}
\mathbf{W}_n = \mathbf{W}_0 + 
\begin{cases}
\displaystyle \sum_{i \in \mathcal{I}_n} \alpha_{n,i}\mathbf{E}_i, & \text{if } \alpha_{n,n} \geq p, \\
\displaystyle \sum\limits_{\substack{i \in \mathcal{I}_n, \\ i \neq n}} \alpha_{n,i}\mathbf{E}_i, & \text{otherwise},
\end{cases}
\label{con:p}
\end{equation}
where ${\alpha}_{n,n}$ is the coefficient assigned to the newly added expert for task $n$. 
This expert pruning strategy can effectively alleviate knowledge redundancy and reduce model parameters.



\begin{figure*}[t]
    \centering  
    \includegraphics[width=0.9\linewidth]{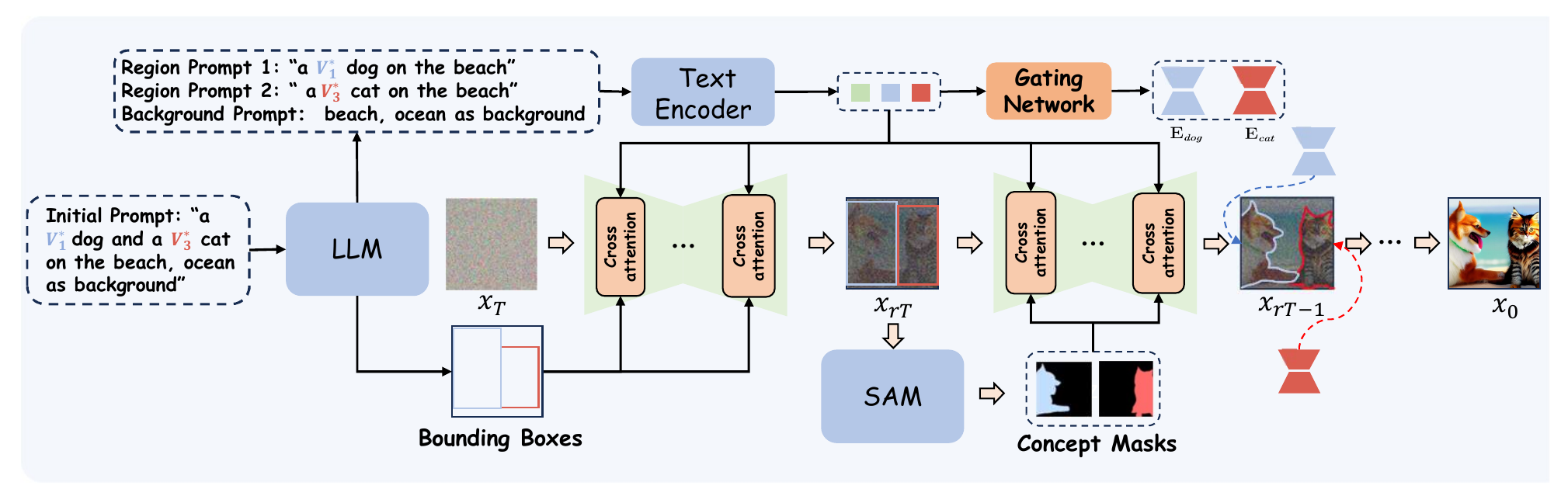}
    \vspace{-10pt}
    \caption{\textbf{Inference pipeline for our proposed R$^2$MoE framework.}
    First, the initial text prompt is entered into GPT-4 to generate the corresponding region prompt and the bounding box.
    For $t>rT$, the model is guided by bounding boxes to control the spatial layout of generated image. 
    For $t\le rT$, the image segmentation model SAM is employed to extract more refined concept masks to further enhance visual fidelity. 
    This two-stage approach mitigates concept overlap and reduces interference between different concepts while better utilizing the generative power of pre-trained diffusion model.
    }
    \label{fig:mul-gen-method}
    \vspace{-10pt}
\end{figure*}

\subsection{Hierarchical Local Attention-Guided Inference}
\label{sec:S4D}

Personalized text-to-image generation models often suffer from concept omission and feature entanglement when integrating newly learned concepts with existing subjects or background elements. These limitations hinder the model's ability to generate images that align with input text prompts. To address these issues, we propose a hierarchical local attention-guided inference approach. Our approach leverages GPT-4~\cite{GPT4} to generate concept-specific bounding boxes and region prompts. It then controls the generation of each concept by computing local attention during early denoising stages and further refines the generation results using concept masks extracted from SAM~\cite {sam} in late denoising stages.

At the inference phase, the model synthesizes an image conditioned on the user-provided text prompt $c$. 
To facilitate fine-grained control over the generation process, we employ GPT-4 to decompose $c$ into a background prompt $c_0$, a set of $\mathcal{U}$ region prompts $\{{c_1, c_2,...,c_{\mathcal{U}}\}}$.
For each region prompt, we also generate a corresponding bounding box
$\textbf{b}_{u}$ (See Fig.\ref{fig:mul-gen-method} for an illustration).
Using these bounding boxes, we construct a set of binary region masks $\mathbf{M} = \{\mathbf{M}_u\in \mathbb{R}^{h\times w} \}^\mathcal{U}_{u=1}$, where values within the bounding box region are set to $1$ and all others to $0$. These masks guide localized attention computation within the cross-attention layers.
For each region, we perform region-specific attention to obtain the corresponding output $\mathbf{A}_u$, formulated as:
\begin{equation}
\mathbf{A}_u=\text{Attention}(\mathbf{Q},\mathbf{K}_u,\mathbf{V}_u)=\text{Softmax}(\frac{\mathbf{Q}\mathbf{K}_u^\top}{\sqrt{d_k}})\mathbf{V}_u,\label{con:Zu}
\end{equation}
where $\mathbf{A}_u$ is the cross-attention output for the $u$-th region. Specifically, $\mathbf{Q} = \mathbf{W}^Q\mathbf{z} \in \mathbb{R}^{h\times w\times d_k}$ is the query matrix computed from the image latent feature $\mathbf{z}$, while the key and value matrices are defined as $\mathbf{K}_u = \mathbf{W}^K_u\mathbf{C}_u \in \mathbb{R}^{d_c\times d_k}$ and $\mathbf{V}_u = \mathbf{W}^V_u\mathbf{C}_u \in \mathbb{R}^{d_c\times d_k}$, respectively. Here, $\mathbf{C}_u$ is the text embedding of the region prompt $c_u$, and the matrices $\mathbf{W}^K_u$ and $\mathbf{W}^V_u$ are the expert-composed attention weights corresponding to the $u$-th concept. 
These are dynamically derived from the expert combination relevant to the region, while $\mathbf{W}^Q$ is shared across all regions.
The modified cross-attention output $\mathbf{\hat{A}}$ is computed as follows:
\begin{equation} 
\mathbf{\hat{A}}= 
(1-\bigcup_{u=1}^{\mathcal{U}} \textbf{M}_{u}) \odot\mathbf{A}_{0}+\gamma\sum_{u=1}^{\mathcal{U}} \textbf{M}_{u} \odot\mathbf{A}_{u},\label{con:hatz2}
\end{equation} 
where $\mathbf{A}_\text{0}$ represents the cross-attention output derived from the background prompt $c_0$ using the pre-trained weight, and $\gamma \in [0,1]$ is a hyperparameter that controls the contribution of the expert-based attention.
When the timestep $t>rT$, we use the bounding boxes to define the region masks with a small $\gamma$, guiding attention toward specific regions to ensure alignment between the generated layout and the text description.
After $rT$ steps, we refine the attention masks by employing SAM to identify the regions most relevant to the semantics of each concept.
These refined masks replace the original bounding box-based masks to achieve further refinement of each concept's generation region.
This hierarchical attention strategy mitigates interference between generated visual concepts while effectively leveraging pre-trained model priors to enhance the consistency of the generated images and text prompts.

\section{Experiments}
This section presents extensive experiments evaluating our proposed $\text{R}^2$ MoE framework. 
We begin with the experimental setup (Sec.\ref{sec:S5A}), followed by qualitative and quantitative comparisons with existing lifelong concept generation methods across multiple benchmarks (Sec.\ref{sec:S5B} and Sec.\ref{sec:S5C}). 
We then conduct ablation studies validating each component's contribution (Sec.\ref{sec:S5D}) and please refer to the supplementary material for more experimental results.

\subsection{Experimental Setup}
\label{sec:S5A}

\noindent\textbf{Datasets.} Our evaluation utilizes the DreamBooth~\cite{dreambooth} (30 subjects) and CustomConcept101~\cite{customdiffusion} (101 concepts) datasets, structured as sequential lifelong learning tasks with randomized ordering. For comparative consistency, we also employ 5-task and 10-task benchmarks aligned with L2DM~\cite{l2dm} and CIDM~\cite{CIDM} respectively. This comprehensive setup ensures robust assessment of lifelong concept learning performance. 

\noindent\textbf{Baselines.}
We benchmark our method against state-of-the-art lifelong concept learning approaches, including CIDM~\cite{CIDM}, L2DM~\cite{l2dm}, and C-LoRA~\cite{clora}. Additionally, we compare with classic continual learning methods (EWC~\cite{ewc}, LwF~\cite{lwf}), personalized generation techniques (CustomDiffusion~\cite{customdiffusion}, LoRA-C~\cite{Multi-LoRA}, OFT~\cite{oft}), and a sequential LoRA fine-tuning baseline. These comparisons collectively demonstrate the advantages of our R$^2$MoE framework over existing methods.

\noindent\textbf{Implementation Details.}
We build upon Stable Diffusion 1.5 as the base model and integrate MoE modules into the Key and Value projection matrices of the cross-attention layers.
During training, only the MoE layers and token embedding layer parameters are updated, with each task trained for 800 iterations.
The learning rates are set to $1\times 10^{-4}$ for the MoE module and $1\times 10^{-3}$ for the text embeddings. 
All models are trained with a batch size of $2$.
During inference, we employ a DDIM sampler~\cite{ddpm} with $50$ steps combined with classifier-free guidance~\cite{ho2022classifier} using a scale of $7$ for all methods..

\noindent\textbf{Evaluation Metrics.} We evaluate lifelong concept learning performance using three key metrics: Image Alignment (IA), measured by CLIP-IA and DINO-IA; Text Alignment (TA); and Forgetting.
Comprehensive experimental configurations and hyperparameter settings are detailed in the supplementary material.
\begin{table*}[ht]
    \centering
    \caption{\textbf{Comparison of our method with state-of-the-art approaches in terms of image alignment, text alignment, and forgetting on the DreamBooth and CustomConcept101 datasets.} The best results are shown in \textbf{bold}, and the second-best results are \underline{underlined}. \textit{Stable Diffusion} denotes the base model without fine-tuning and serves as a reference.}
    \resizebox{1.0\linewidth}{!}{
    \begin{tabular}{l|cccc|cccc}
        \toprule
        \multirow{2}{*}{\textbf{Comparison Methods}} & \multicolumn{4}{c|}{\textbf{DreamBooth}} & \multicolumn{4}{c}{\textbf{CustomConcept101}} \\
        & {$\text{CLIP-IA}(\uparrow)$} & {$\text{CLIP-TA}(\uparrow)$}& {$\text{DINO-IA}(\uparrow)$} & {$\text{Forgetting}(\downarrow)$} & {$\text{CLIP-IA}(\uparrow)$} & {$\text{CLIP-TA}(\uparrow)$} & {$\text{DINO-IA}(\uparrow)$} &{$\text{Forgetting}(\downarrow)$} \\
        \midrule
         \textit{Stable Diffusion} &64.1&79.6&27.8 & - &61.9 & 79.5&25.5&-\\
         \midrule
         LoRA\cite{lora} &69.9 & 70.6& 54.3 & 7.18 &66.9 &62.3 &27.6 &9.13  \\
         OFT \cite{oft} &70.1 & 71.5& 41.7 & 5.91 &67.5 &69.4 &33.5&5.53  \\
         C-LoRA \cite{clora} &76.5&72.6&58.2&1.12& 72.5&72.5&42.6&1.56   \\
          CIDM \cite{CIDM} &67.3&\textbf{77.9}&38.5&8.03& 65.2&\textbf{76.3}&33.2&8.56   \\
          \midrule
         \rowcolor{ours}\textbf{Ours w/o HLAG} & \textbf{78.8}&71.0 &\textbf{62.0} &\underline{0.06} &\underline{73.6}&75.5 &\underline{46.5}&\underline{0.23} \\
         
         \rowcolor{ours}\textbf{Ours} &\underline{78.7}& \underline{73.5} &\underline{61.9} &\textbf{0.05} & \textbf{75.3}& \underline{76.0}& \textbf{50.5}& \textbf{0.19}  \\
         \bottomrule
    \end{tabular}
    }
    \label{tab:evaluation_on_sota}
\end{table*}

\subsection{Image Generation Quality Comparison}
\label{sec:S5B}

\begin{table}[ht]
\centering
\vspace{0pt}
\caption{\textbf{Comparison of image alignment, text alignment, and forgetting between our method and state-of-the-art approaches on the 5-task benchmark.}}
\resizebox{\linewidth}{!}{
\begin{tabular}{l|ccc}
\toprule
\textbf{ Comparison Methods} & CLIP-IA($\uparrow$) & CLIP-TA($\uparrow$) & Forgetting($\downarrow$) \\  
\midrule
Custom~\cite{customdiffusion}     & 77.8 & 64.8 & 4.3  \\ 
Custom~\cite{customdiffusion}+EWC~\cite{ewc}   & 78.1 & 65.3 & 3.6   \\
Custom~\cite{customdiffusion}+LwF~\cite{lwf}  & 76.1 & 65.3 & 4.6   \\
OFT~\cite{oft} &75.7 &\textbf{76.2} &4.2  \\
L2DM~\cite{l2dm}  & 80.4 & 75.0 & 1.2    \\
C-LoRA~\cite{clora} & 82.7&73.7 &0.6 \\
CIDM~\cite{CIDM}  & 84.1 & 74.3&\underline{0.4}  \\ 
\midrule
\rowcolor{ours}\textbf{Ours w/o HLAG}   &\textbf{85.7} &73.5&\textbf{0.01} \\
\rowcolor{ours}\textbf{Ours}  &\underline{84.3} &\underline{75.4} &\textbf{0.01}  \\ 
\bottomrule
\end{tabular}
}
\label{table:ta_five}
    \vspace{0pt}
\end{table}

\begin{table}[ht]
\centering
\vspace{0pt}
\caption{\textbf{Comparison of image alignment, text alignment, and forgetting between our method and state-of-the-art approaches on the 10-task benchmark.}}
\resizebox{\linewidth}{!}{
\begin{tabular}{l|ccc}
\toprule
\textbf{ Comparison Methods} & CLIP-IA($\uparrow$) & CLIP-TA($\uparrow$) & Forgetting($\downarrow$) \\  	 	 	 
\midrule
EWC~\cite{ewc}   &75.9  &72.7  &2.48   \\
LwF~\cite{lwf}  & 74.1 &73.4  &   -\\

LoRA-M~\cite{Multi-LoRA} &74.6 & -&4.42 \\
LoRA-C~\cite{Multi-LoRA}  &74.9 & -&4.30  \\ 
C-LoRA~\cite{clora}  & 76.9&73.6  &1.56   \\
L2DM~\cite{l2dm}  & 76.1& 72.7 &1.37   \\
CIDM~\cite{CIDM}  &78.0 &\underline{74.8} &1.22   \\
\midrule
\rowcolor{ours}\textbf{Ours w/o HLAG}   &\textbf{82.3} &72.1&\textbf{0.02} \\
\rowcolor{ours}\textbf{Ours}  &\underline{80.1} &\textbf{74.9} &\textbf{0.02}  \\ 
\bottomrule
\end{tabular}
}
\label{table:ta_ten}
    \vspace{-10pt}
\end{table}

\noindent\textbf{Quantitative Results.}
To validate the superior lifelong learning capabilities of our framework, we present results in
Tab~\ref{tab:evaluation_on_sota}, which demonstrate that our R$^2$MoE outperforms existing methods on both DreamBooth and CustomConcept101 datasets.
In terms of visual fidelity, our method achieves the highest scores in both CLIP and DINO image alignment metrics. Specifically, on the DreamBooth dataset, we achieve relative improvements of $3.0\%$ and $6.5\%$ in DINO-IA and CLIP-IA, respectively. On the CustomConcept101 dataset, the corresponding improvements reach $3.9\%$ and $18.5\%$, respectively. Regarding knowledge retention, our method demonstrates the lowest forgetting rates: $0.05\%$ on DreamBooth and $0.19\%$ on CustomConcept101, indicating strong resistance to catastrophic forgetting.
In terms of text alignment, R$^2$MoE demonstrates strong competitiveness, outperforming most existing methods.

In contrast to our approach, existing methods exhibit significant catastrophic forgetting in lifelong concept learning scenarios.
C-LoRA, which learns all tasks within a shared low-rank space, experiences gradient conflicts as the number of tasks increases, resulting in higher forgetting rates. 
Since CIDM's elastic weight consolidation module relies on input prompt semantics to select relevant LoRA components, when learning numerous semantically similar concepts, this module struggles to distinguish between new and existing concepts, leading to catastrophic forgetting. 
The substantial improvements in image alignment and reduced forgetting rates validate R$^2$MoE's core design: expert modules effectively acquire new concept knowledge while the gating network maintains stable routing capabilities without significant forgetting.
\begin{figure*}[t!]
    \centering  
    \includegraphics[width=0.95\linewidth]{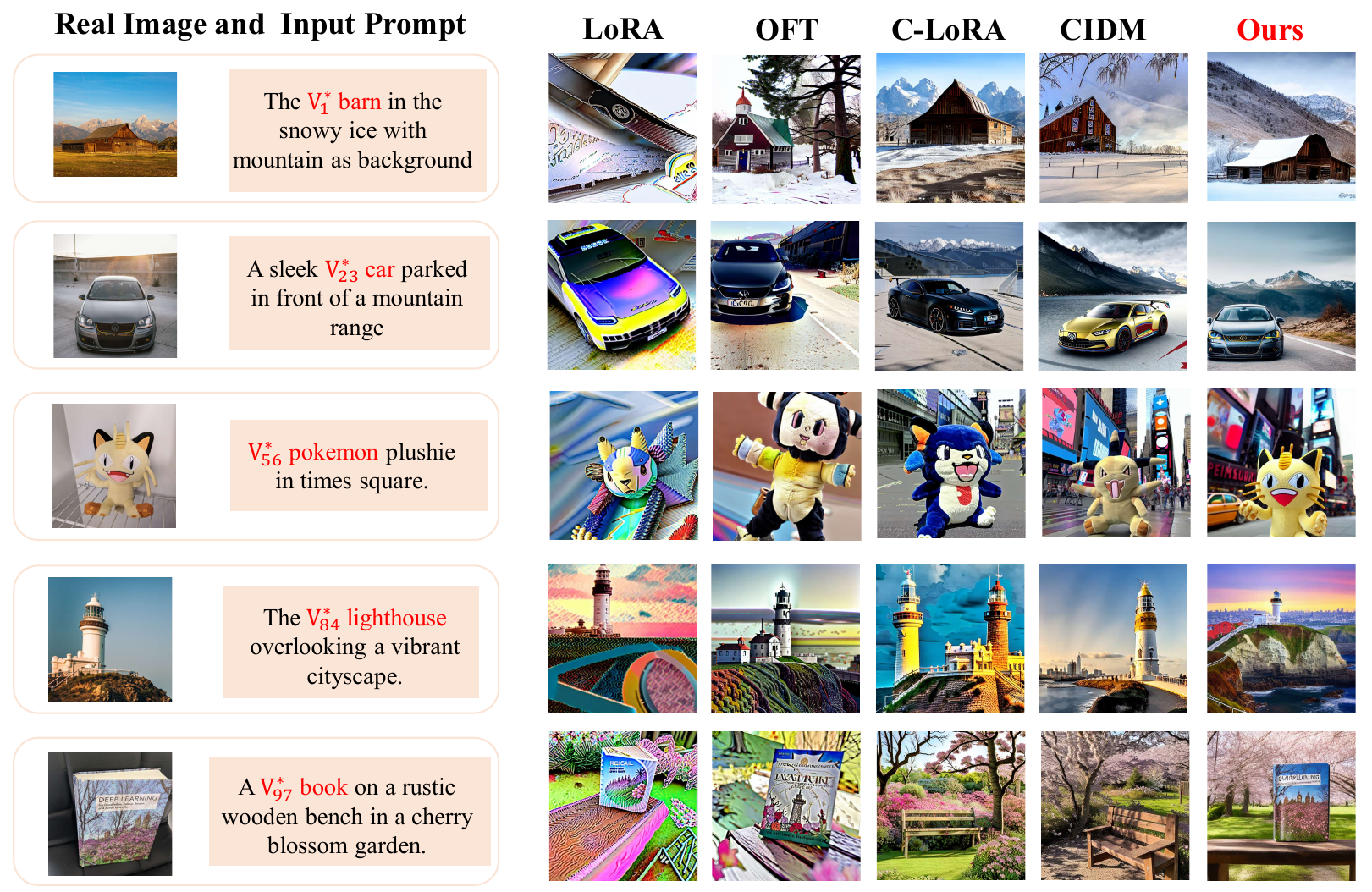}
    \caption{\textbf{Qualitative comparison with existing state-of-the-art customization methods.} 
    R$^2$MoE produces higher quality results with better subject similarity and text controllability compared to existing methods.
    For instance, our approach successfully generates images for the first task containing elements(ice, mountain and barn), while alternative methods either fail to accurately represent the concept or omit critical elements like mountains and ice, demonstrating our method's lower forgetting rate. 
    The results in the final row indicate that only our method successfully generates images incorporating book, bench and garden elements, evidencing the superior generalization capability of our approach.}
    \label{fig:Qualitative results}
    \vspace{-8pt}
\end{figure*}
To demonstrate the effectiveness of our framework in short task sequences, we evaluate R$^2$MoE on both 5-task and 10-task benchmarks.
Tab~\ref{table:ta_five} shows the result on the 5-task benchmark, with $0.2\%,\,1.5\%$ improvements in image and text alignment respectively over CIDM, while sustaining a low forgetting rate of $0.01\%$.
Tab~\ref{table:ta_ten} presents the comparative results on the 10-task benchmark, where our method outperforms previous SOTA methods in terms of image alignment, text alignment, and forgetting rate.
Compared to CIDM, our approach improves of $2.7\%,\,0.2\%$ in image alignment and text alignment respectively, while maintaining a minimal forgetting rate of $0.02\%$.
These results demonstrate the proposed R$^2$MoE effectiveness even in shorter task sequences.

\noindent\textbf{Qualitative Results.}
Fig.\ref{fig:Qualitative results} illustrates the qualitative results on the CustomConcept101 dataset.
Results generated by LoRA and OFT show pronounced visual degradation, indicating that continuous fusion of parameters from different tasks leads to catastrophic forgetting.
Although C-LoRA and CIDM can generate images consistent with text descriptions, the poor conceptual fidelity of the generated images indicates that both methods still exhibit obvious catastrophic forgetting.
Furthermore, these methods exhibit poor plasticity in long task sequences. 
For example, they can generate the earlier-learned concept "barn" (first line), but fail to generate the later-learned concept "book" (last line).
Compared with the above methods, our R$^2$MoE demonstrates excellent visual fidelity and semantic alignment across all tasks, indicating its strong lifelong concept learning ability.

\begin{figure*}[t!]
    \centering  
    \includegraphics[width=0.95\linewidth]{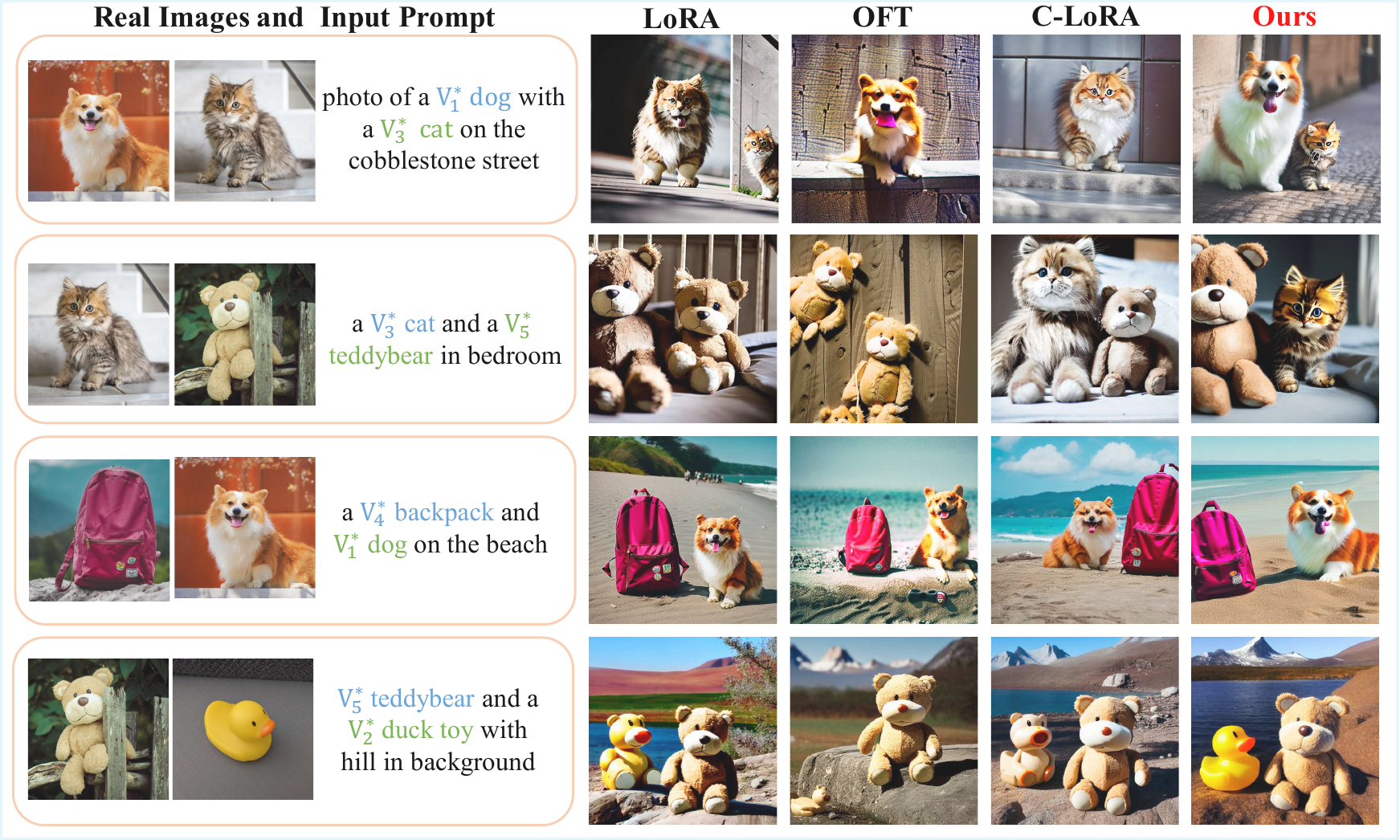}
    \vspace{-10pt}
    \caption{\textbf{Qualitative Comparison for Multi-object Generation.}
    We demonstrate that R$^2$MoE  generates multi-subject customized results with superior text controllability and subject fidelity compared to existing lifelong learning methods. 
    For instance, in the first row, R$^2$MoE successfully generates  specific cat and dog in street scenes, whereas alternative approaches suffer from concept omission and feature entanglement issues.
    }
    \label{fig:Qualitative results multi}
    \vspace{-10pt}
\end{figure*}

As evidenced in Fig.\ref{fig:Qualitative results multi}, our approach achieves superior multi-concept generation performance. Whereas competing methods commonly encounter concept omission and feature entanglement problems, our framework effectively resolves these limitations, synthesizing images that successfully incorporate multiple personalized concepts with heightened visual fidelity and semantic coherence. 
This indicates that our method retains the pre-trained diffusion model's core generative capabilities while facilitating precise compositional control, establishing an effective equilibrium between generalization and specificity that previous approaches have failed to achieve.
\begin{table}[htb] {  
    \centering  
    \caption{\textbf{Comparison of parameter with alternative methods across different task lengths.}
    Imp. indicates the the drop ratio of our method compared to that method.}  
    \resizebox{1.0\linewidth}{!}{\begin{tabular}{l|cc|cc}  
        \toprule  
        \multirow{2}{*}{\textbf{Model/Method}} &  \multicolumn{2}{c|}{\textbf{30 Tasks}} & \multicolumn{2}{c}{\textbf{101 Tasks}} \\  
        \cmidrule(lr){2-3} \cmidrule(lr){4-5}  
        & Total Params &Imp. & Total Params & Imp.  \\  
        \midrule  
        \textbf{Pre-trained Model} & 860M & - & 860M & - \\
        \midrule  
        OFT \cite{oft} & +667.8M & \rotatebox[origin=c]{180}{$\Uparrow$}99.4\% & +2248.3M &\rotatebox[origin=c]{180}{$\Uparrow$}99.3\% \\   
        CustomDiffusion \cite{customdiffusion}& +18.3M & \rotatebox[origin=c]{180}{$\Uparrow$}77.0\% & +18.3M &\rotatebox[origin=c]{180}{$\Uparrow$}16.9\% \\
        C-LoRA \cite{clora} & +24.0M &\rotatebox[origin=c]{180}{$\Uparrow$}82.5\% & +80.8M &\rotatebox[origin=c]{180}{$\Uparrow$}81.2\% \\   
        CIDM \cite{CIDM} & +12.3M &\rotatebox[origin=c]{180}{$\Uparrow$}65.9\% & +41.4M &\rotatebox[origin=c]{180}{$\Uparrow$}63.3\% \\  
        \midrule   
        \rowcolor{ours}\textbf{Our w/o LRER} & +6.8M &\rotatebox[origin=c]{180}{$\Uparrow$}38.2\% & +22.9M &\rotatebox[origin=c]{180}{$\Uparrow$}33.6\% \\
        \rowcolor{ours}\textbf{Our} & \textbf{+4.2M} & - & \textbf{+15.2M} & - \\
        \bottomrule  
    \end{tabular}}  
    \label{tab:Storge Params}  
    }   
    \vspace{-10pt}
\end{table}

\subsection{Model Size Comparison} 
\label{sec:S5C}
Tab~\ref{tab:Storge Params} demonstrates the superiority of $\text{R}^2$MoE in terms of model size. 
CustomDiffusion stores cross-attention modules of the diffusion model, resulting in a relatively large number of additional parameters. OFT introduces high-dimensional orthogonal matrices for each concept (reaching 22.3M parameters per concept), yielding the highest storage cost among all methods. 
While C-LoRA and CIDM reduce parameter usage through low-rank adaptation matrices, they fail to effectively address parameter redundancy accumulation during continual learning. 
Our method incorporates a layer-wise redundant expert removal (LRER) strategy. that automatically eliminates underutilized experts after each task, effectively addressing parameter redundancy. 
This leads to a $38.2\%$ reduction in model parameters on the DreamBooth dataset and a $33.6\%$ reduction on the CustomConcept101 dataset compared to the baselines without pruning.
Compared to second-best baseline CIDM, $\text{R}^2$MoE achieves storage efficiency improvements of $65.9\%$ and $63.3\%$ in DreamBooth and CustomConcept101 benchmarks, respectively.
\begin{table}[ht]
    \centering
    \caption{\textbf{Quantitative ablation study of key components.} }
    \adjustbox{max width=\linewidth}{  
    \begin{tabular}{l|cc}
        \toprule
        \multirow{2}{*}{\textbf{Comparison Methods}}
        &\multicolumn{2}{c}{\textbf{DreamBooth}} \\
        &$\text{CLIP-IA}(\uparrow)$ & $\text{DINO-IA}(\uparrow)$ \\
        \midrule
        R$^2$MoE &78.8&62.0  \\
        w/o RDM &75.8&55.5 \\
        w/o RDM \& SAE &73.5&47.9  \\
        
        \bottomrule
    \end{tabular}
    }
    \label{tab:ablation_study}
\end{table}

\subsection{Ablation Study}
\label{sec:S5D}

\noindent\textbf{Component-wise Ablation Analysis.}
Tab~\ref{tab:ablation_study} presents a ablation analysis conducted on the DreamBooth benchmark to evaluate critical components of our framework.
Removing the Routing Distillation Mechanism (RDM) results in a significant drop in both CLIP and DINO image alignment scores (15.3\% and 17.8\% reduction, respectively).
This is because overwriting the gating network parameters destroys its routing capability to previous tasks.
When both the RDM and Selectively Activating Experts (SAE) strategies are removed, our $\text{R}^2$MoE tends to activate previously trained experts rather than activating new ones. 
This causes the model to degrade into a single-expert mechanism, severely hindering its ability to learn new concepts.

\begin{figure}[htb]
    \centering  
    \includegraphics[width=0.95\linewidth]{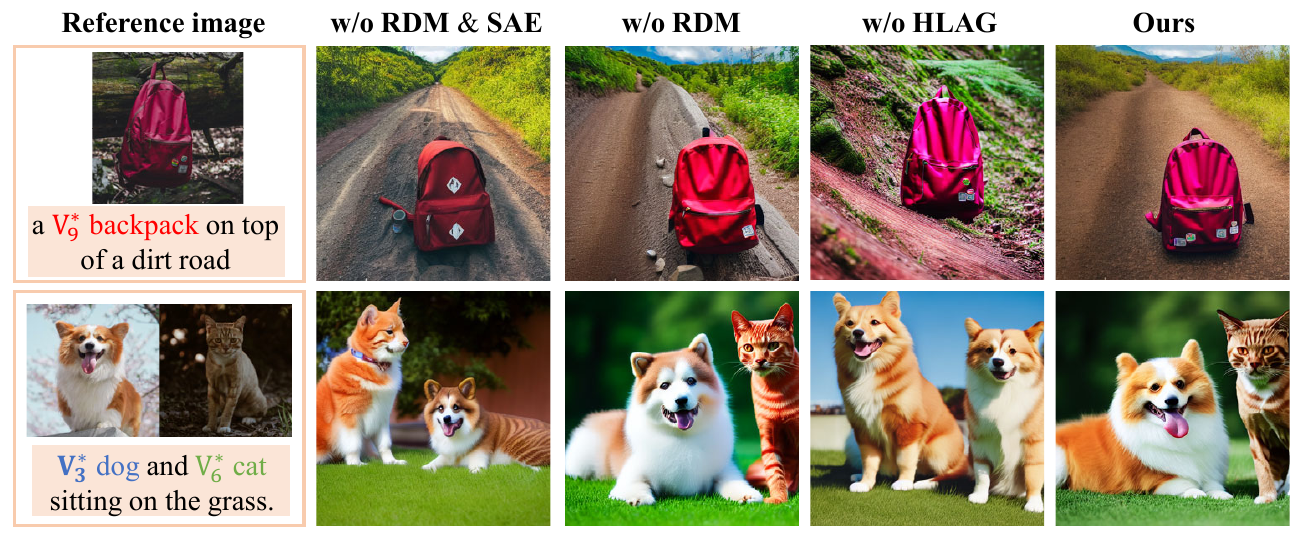}\\
    \caption{\textbf{Qualitative ablation study.}
    The first row presents the ablation results for single-concept generation, while the second row shows the results for multi-concept generation.
    }
    \label{fig:ablation_study}
    \vspace{-10pt}
\end{figure}

Fig.\ref{fig:ablation_study} illustrates the impact of ablating each key component on generation results. 
The top row displays single-concept generation results. 
Removing the HLAG module causes the generation of overfitted training artifacts (e.g., background weeds from training samples), compromising semantic alignment with input prompts. 
Removing RDM  significantly reduces concept visual fidelity, as evidenced by the missing badge on the backpack. 
Eliminating both RDM and the SAE module further deteriorates visual quality. 
The second row of Fig.\ref{fig:ablation_study} shows multi-concept generation results. 
Without the HLAG module, generated images exhibit clear feature entanglement and concept omission, producing two dogs with similar features while failing to generate the cat and grass, resulting in decreased visual fidelity. Removing RDM causes significant concept fidelity degradation, such as the dog's fur incorrectly changing to white.

\section{Conclusion}
In this paper, we propose R$^2$MoE, a novel Mixture-of-Experts framework that tackles the dual challenges of catastrophic forgetting and parameter expansion in personalized lifelong concept learning. 
Our approach employs routing distillation to preserve task-specific knowledge while systematically eliminating redundant experts to minimize parameter overhead. Additionally, we propose a hierarchical local attention-guided inference strategy that enhances concept disentanglement and reduces interference between learned visual concepts. Comprehensive experiments validate that R²MoE achieves superior performance compared to state-of-the-art methods while maintaining computational efficiency, demonstrating its practical value for lifelong learning applications.

\section*{Acknowledgement}
This work is supported by the National Natural Science Foundation of China (No.~62171111).

\bibliographystyle{IEEEtran}
\bibliography{TMM2024}

\begin{thebibliography}{10}
\providecommand{\url}[1]{#1}
\csname url@samestyle\endcsname
\providecommand{\newblock}{\relax}
\providecommand{\bibinfo}[2]{#2}
\providecommand{\BIBentrySTDinterwordspacing}{\spaceskip=0pt\relax}
\providecommand{\BIBentryALTinterwordstretchfactor}{4}
\providecommand{\BIBentryALTinterwordspacing}{\spaceskip=\fontdimen2\font plus
\BIBentryALTinterwordstretchfactor\fontdimen3\font minus \fontdimen4\font\relax}
\providecommand{\BIBforeignlanguage}[2]{{%
\expandafter\ifx\csname l@#1\endcsname\relax
\typeout{** WARNING: IEEEtran.bst: No hyphenation pattern has been}%
\typeout{** loaded for the language `#1'. Using the pattern for}%
\typeout{** the default language instead.}%
\else
\language=\csname l@#1\endcsname
\fi
#2}}
\providecommand{\BIBdecl}{\relax}
\BIBdecl

\bibitem{sd}
R.~Rombach, A.~Blattmann, D.~Lorenz, P.~Esser, and B.~Ommer, ``High-resolution image synthesis with latent diffusion models,'' in \emph{Proceedings of the IEEE/CVF conference on computer vision and pattern recognition}, 2022, pp. 10\,684--10\,695.

\bibitem{sdxl}
D.~Podell, Z.~English, K.~Lacey, A.~Blattmann, T.~Dockhorn, J.~M{\"u}ller, J.~Penna, and R.~Rombach, ``Sdxl: Improving latent diffusion models for high-resolution image synthesis,'' \emph{arXiv preprint arXiv:2307.01952}, 2023.

\bibitem{sd3}
P.~Esser, S.~Kulal, A.~Blattmann, R.~Entezari, J.~M{\"u}ller, H.~Saini, Y.~Levi, D.~Lorenz, A.~Sauer, F.~Boesel \emph{et~al.}, ``Scaling rectified flow transformers for high-resolution image synthesis,'' in \emph{Forty-first international conference on machine learning}, 2024.

\bibitem{Imagen}
C.~Saharia, W.~Chan, S.~Saxena, L.~Li, J.~Whang, E.~L. Denton, K.~Ghasemipour, R.~Gontijo~Lopes, B.~Karagol~Ayan, T.~Salimans \emph{et~al.}, ``Photorealistic text-to-image diffusion models with deep language understanding,'' \emph{Advances in neural information processing systems}, vol.~35, pp. 36\,479--36\,494, 2022.

\bibitem{ddpm}
J.~Ho, A.~Jain, and P.~Abbeel, ``Denoising diffusion probabilistic models,'' \emph{Advances in neural information processing systems}, vol.~33, pp. 6840--6851, 2020.

\bibitem{textureinversion}
R.~Gal, Y.~Alaluf, Y.~Atzmon, O.~Patashnik, A.~H. Bermano, G.~Chechik, and D.~Cohen-Or, ``An image is worth one word: Personalizing text-to-image generation using textual inversion,'' \emph{arXiv preprint arXiv:2208.01618}, 2022.

\bibitem{dreambooth}
N.~Ruiz, Y.~Li, V.~Jampani, Y.~Pritch, M.~Rubinstein, and K.~Aberman, ``Dreambooth: Fine tuning text-to-image diffusion models for subject-driven generation,'' in \emph{Proceedings of the IEEE/CVF Conference on Computer Vision and Pattern Recognition}, 2023, pp. 22\,500--22\,510.

\bibitem{oft}
Z.~Qiu, W.~Liu, H.~Feng, Y.~Xue, Y.~Feng, Z.~Liu, D.~Zhang, A.~Weller, and B.~Sch{\"o}lkopf, ``Controlling text-to-image diffusion by orthogonal finetuning,'' \emph{Advances in Neural Information Processing Systems}, vol.~36, pp. 79\,320--79\,362, 2023.

\bibitem{Personalized-Residuals}
C.~Ham, M.~Fisher, J.~Hays, N.~Kolkin, Y.~Liu, R.~Zhang, and T.~Hinz, ``Personalized residuals for concept-driven text-to-image generation,'' in \emph{Proceedings of the IEEE/CVF Conference on Computer Vision and Pattern Recognition}, 2024, pp. 8186--8195.

\bibitem{xu2024sgdm}
Y.~Xu, X.~Xu, H.~Gao, and F.~Xiao, ``Sgdm: an adaptive style-guided diffusion model for personalized text to image generation,'' \emph{IEEE Transactions on Multimedia}, vol.~26, pp. 9804--9813, 2024.

\bibitem{zhang2024two}
S.~Zhang, M.~Ni, S.~Chen, L.~Wang, W.~Ding, and Y.~Liu, ``A two-stage personalized virtual try-on framework with shape control and texture guidance,'' \emph{IEEE Transactions on Multimedia}, vol.~26, pp. 10\,225--10\,236, 2024.

\bibitem{chen2025videodreamer}
H.~Chen, X.~Wang, G.~Zeng, Y.~Zhang, Y.~Zhou, F.~Han, Y.~Wu, and W.~Zhu, ``Videodreamer: Customized multi-subject text-to-video generation with disen-mix finetuning on language-video foundation models,'' \emph{IEEE Transactions on Multimedia}, 2025.

\bibitem{customdiffusion}
N.~Kumari, B.~Zhang, R.~Zhang, E.~Shechtman, and J.-Y. Zhu, ``Multi-concept customization of text-to-image diffusion,'' in \emph{Proceedings of the IEEE/CVF Conference on Computer Vision and Pattern Recognition}, 2023, pp. 1931--1941.

\bibitem{han2023svdiff}
L.~Han, Y.~Li, H.~Zhang, P.~Milanfar, D.~Metaxas, and F.~Yang, ``Svdiff: Compact parameter space for diffusion fine-tuning,'' \emph{arXiv preprint arXiv:2303.11305}, 2023.

\bibitem{animediff}
Y.~Jiang, Q.~Liu, D.~Chen, L.~Yuan, and Y.~Fu, ``Animediff: Customized image generation of anime characters using diffusion model,'' \emph{IEEE Transactions on Multimedia}, 2024.

\bibitem{mixofshow}
Y.~Gu, X.~Wang, J.~Z. Wu, Y.~Shi, Y.~Chen, Z.~Fan, W.~Xiao, R.~Zhao, S.~Chang, W.~Wu \emph{et~al.}, ``Mix-of-show: Decentralized low-rank adaptation for multi-concept customization of diffusion models,'' \emph{Advances in Neural Information Processing Systems}, vol.~36, 2024.

\bibitem{cvpr2024orthogonal}
R.~Po, G.~Yang, K.~Aberman, and G.~Wetzstein, ``Orthogonal adaptation for modular customization of diffusion models,'' in \emph{Proceedings of the IEEE/CVF Conference on Computer Vision and Pattern Recognition}, 2024, pp. 7964--7973.

\bibitem{Multi-LoRA}
M.~Zhong, Y.~Shen, S.~Wang, Y.~Lu, Y.~Jiao, S.~Ouyang, D.~Yu, J.~Han, and W.~Chen, ``Multi-lora composition for image generation,'' \emph{CoRR}, 2024.

\bibitem{zhang2025multi}
H.~Zhang, T.~Wu, and Y.~Wei, ``Multi-view user preference modeling for personalized text-to-image generation,'' \emph{IEEE Transactions on Multimedia}, 2025.

\bibitem{clora}
J.~S. Smith, Y.-C. Hsu, L.~Zhang, T.~Hua, Z.~Kira, Y.~Shen, and H.~Jin, ``Continual diffusion: Continual customization of text-to-image diffusion with c-lora,'' \emph{Transactions on Machine Learning Research}, 2024.

\bibitem{l2dm}
G.~Sun, W.~Liang, J.~Dong, J.~Li, Z.~Ding, and Y.~Cong, ``Create your world: Lifelong text-to-image diffusion,'' \emph{IEEE Transactions on Pattern Analysis and Machine Intelligence}, 2024.

\bibitem{CIDM}
J.~Dong, W.~Liang, H.~Li, D.~Zhang, M.~Cao, H.~Ding, S.~H. Khan, and F.~Shahbaz~Khan, ``How to continually adapt text-to-image diffusion models for flexible customization?'' \emph{Advances in Neural Information Processing Systems}, vol.~37, pp. 130\,057--130\,083, 2024.

\bibitem{guo2025conceptguard}
Z.~Guo and T.~Jin, ``Conceptguard: Continual personalized text-to-image generation with forgetting and confusion mitigation,'' in \emph{Proceedings of the Computer Vision and Pattern Recognition Conference}, 2025, pp. 2945--2954.

\bibitem{mccloskey1989catastrophic}
M.~McCloskey and N.~J. Cohen, ``Catastrophic interference in connectionist networks: The sequential learning problem,'' in \emph{Psychology of learning and motivation}.\hskip 1em plus 0.5em minus 0.4em\relax Elsevier, 1989, vol.~24, pp. 109--165.

\bibitem{GPT4}
J.~Achiam, S.~Adler, S.~Agarwal, L.~Ahmad, I.~Akkaya, F.~L. Aleman, D.~Almeida, J.~Altenschmidt, S.~Altman, S.~Anadkat \emph{et~al.}, ``Gpt-4 technical report,'' \emph{arXiv preprint arXiv:2303.08774}, 2023.

\bibitem{sam}
A.~Kirillov, E.~Mintun, N.~Ravi, H.~Mao, C.~Rolland, L.~Gustafson, T.~Xiao, S.~Whitehead, A.~C. Berg, W.-Y. Lo, P.~Doll{\'a}r, and R.~Girshick, ``Segment anything,'' \emph{arXiv:2304.02643}, 2023.

\bibitem{instantbooth}
J.~Shi, W.~Xiong, Z.~Lin, and H.~J. Jung, ``Instantbooth: Personalized text-to-image generation without test-time finetuning,'' \emph{arXiv preprint arXiv:2304.03411}, 2023.

\bibitem{DALLE-2}
A.~Ramesh, P.~Dhariwal, A.~Nichol, C.~Chu, and M.~Chen, ``Hierarchical text-conditional image generation with clip latents,'' \emph{arXiv preprint arXiv:2204.06125}, vol.~1, no.~2, p.~3, 2022.

\bibitem{lora}
E.~J. Hu, Y.~Shen, P.~Wallis, Z.~Allen-Zhu, Y.~Li, S.~Wang, L.~Wang, and W.~Chen, ``Lora: Low-rank adaptation of large language models,'' \emph{arXiv preprint arXiv:2106.09685}, 2021.

\bibitem{moe}
R.~A. Jacobs, M.~I. Jordan, S.~J. Nowlan, and G.~E. Hinton, ``Adaptive mixtures of local experts,'' \emph{Neural computation}, vol.~3, no.~1, pp. 79--87, 1991.

\bibitem{sparsemoe}
\BIBentryALTinterwordspacing
N.~Shazeer, A.~Mirhoseini, K.~Maziarz, A.~Davis, Q.~Le, G.~Hinton, and J.~Dean, ``Outrageously large neural networks: The sparsely-gated mixture-of-experts layer,'' 2017. [Online]. Available: \url{https://arxiv.org/abs/1701.06538}
\BIBentrySTDinterwordspacing

\bibitem{MOA}
W.~Feng, C.~Hao, Y.~Zhang, Y.~Han, and H.~Wang, ``Mixture-of-loras: An efficient multitask tuning for large language models,'' \emph{arXiv preprint arXiv:2403.03432}, 2024.

\bibitem{MOLE}
X.~Wu, S.~Huang, and F.~Wei, ``Mixture of lora experts,'' \emph{arXiv preprint arXiv:2404.13628}, 2024.

\bibitem{Expert-gate}
R.~Aljundi, P.~Chakravarty, and T.~Tuytelaars, ``Expert gate: Lifelong learning with a network of experts,'' in \emph{Proceedings of the IEEE conference on computer vision and pattern recognition}, 2017, pp. 3366--3375.

\bibitem{Lifelong-MoE}
W.~Chen, Y.~Zhou, N.~Du, Y.~Huang, J.~Laudon, Z.~Chen, and C.~Cui, ``Lifelong language pretraining with distribution-specialized experts,'' in \emph{International Conference on Machine Learning}.\hskip 1em plus 0.5em minus 0.4em\relax PMLR, 2023, pp. 5383--5395.

\bibitem{boosting-MOE}
J.~Yu, Y.~Zhuge, L.~Zhang, P.~Hu, D.~Wang, H.~Lu, and Y.~He, ``Boosting continual learning of vision-language models via mixture-of-experts adapters,'' in \emph{Proceedings of the IEEE/CVF Conference on Computer Vision and Pattern Recognition}, 2024, pp. 23\,219--23\,230.

\bibitem{liveedit}
Q.~Chen, C.~Wang, D.~Wang, T.~Zhang, W.~Li, and X.~He, ``Lifelong knowledge editing for vision language models with low-rank mixture-of-experts,'' \emph{arXiv preprint arXiv:2411.15432}, 2024.

\bibitem{coin}
C.~Chen, J.~Zhu, X.~Luo, H.~Shen, J.~Song, and L.~Gao, ``Coin: A benchmark of continual instruction tuning for multimodel large language models,'' \emph{Advances in Neural Information Processing Systems}, 2024.

\bibitem{dai2024deepseekmoe}
D.~Dai, C.~Deng, C.~Zhao, R.~Xu, H.~Gao, D.~Chen, J.~Li, W.~Zeng, X.~Yu \emph{et~al.}, ``Deepseekmoe: Towards ultimate expert specialization in mixture-of-experts language models,'' \emph{arXiv preprint arXiv:2401.06066}, 2024.

\bibitem{matte}
A.~Agarwal, S.~Karanam, T.~Shukla, and B.~V. Srinivasan, ``An image is worth multiple words: Multi-attribute inversion for constrained text-to-image synthesis,'' in \emph{2025 IEEE/CVF Winter Conference on Applications of Computer Vision (WACV)}.\hskip 1em plus 0.5em minus 0.4em\relax IEEE, 2025, pp. 6053--6062.

\bibitem{ewc}
J.~Kirkpatrick, R.~Pascanu, N.~Rabinowitz, J.~Veness, G.~Desjardins, A.~A. Rusu, K.~Milan, J.~Quan, T.~Ramalho, A.~Grabska-Barwinska \emph{et~al.}, ``Overcoming catastrophic forgetting in neural networks,'' \emph{Proceedings of the national academy of sciences}, vol. 114, no.~13, pp. 3521--3526, 2017.

\bibitem{lwf}
Z.~Li and D.~Hoiem, ``Learning without forgetting,'' \emph{IEEE transactions on pattern analysis and machine intelligence}, vol.~40, no.~12, pp. 2935--2947, 2017.

\bibitem{ho2022classifier}
J.~Ho and T.~Salimans, ``Classifier-free diffusion guidance,'' \emph{arXiv preprint arXiv:2207.12598}, 2022.

\end{thebibliography}

\end{document}